\pdfoutput=1

\documentclass[11pt]{article}

\usepackage{acl}

\usepackage{times}
\usepackage{latexsym}
\usepackage{booktabs}
\usepackage{graphicx}
\usepackage{svg}
\usepackage{amsmath}
\usepackage[T1]{fontenc}

\usepackage[utf8]{inputenc}

\usepackage{microtype}

\title{LIME: Weakly-Supervised Text Classification Without Seeds}

\author{Seongmin Park \and Jihwa Lee \\
         ActionPower, Seoul, Republic of Korea\\
         \texttt{\{seongmin.park, jihwa.lee\}@actionpower.kr}}

\begin{document}
\maketitle
\begin{abstract}
In weakly-supervised text classification, only label names act as sources of supervision. Predominant approaches to weakly-supervised text classification utilize a two-phase framework, where test samples are first assigned pseudo-labels and are then used to train a neural text classifier. In most previous work, the pseudo-labeling step is dependent on obtaining seed words that best capture the relevance of each class label. We present LIME\footnote{\textbf{L}abels \textbf{I}dentified with \textbf{M}aximal \textbf{E}ntailment}, a framework for weakly-supervised text classification that entirely replaces the brittle seed-word generation process with entailment-based pseudo-classification. We find that combining weakly-supervised classification and textual entailment mitigates shortcomings of both, resulting in a more streamlined and effective classification pipeline. With just an off-the-shelf textual entailment model, LIME outperforms recent baselines in weakly-supervised text classification and achieves state-of-the-art in 4 benchmarks. We open source our code at \url{https://github.com/seongminp/LIME}.
\end{abstract}

\section{Introduction}
Weakly-supervised text classification \cite{meng2018weakly} is an important avenue of research in low-resourced text classification. Unlike in traditional text classification, all supervision derives from textual information in category names. Weakly-supervised classification offers a practical approach to classification because it does not necessitate massive amounts of training data.  

Another distinct aspect of weakly-supervised text classification is that the system has access to the entire test set at evaluation time, instead of encountering test samples sequentially. Exploiting this characteristic, recent approaches employ keyword-matching pseudo-labeling schemes to tentatively assign class labels to each test sample, before using the information to train a separate classifier \cite{meng2018weakly, mekala-shang-2020-contextualized, wang-etal-2021-x}. Pseudo-labels are assigned by counting how many “seed words” of each class are found in the test sample. Keyword matching-based labeling, however, is neither adaptable nor flexible because semantic information embedded in class names cannot be extracted adaptively for distinct classification tasks.  

Inspired by recent advances in prompt-based text classification \cite{yin-etal-2019-benchmarking, yin-etal-2020-universal, schick-schutze-2021-exploiting}, we replace the keyword-based pseudo-labeling step with a more streamlined entailment-based approach. Extensive experiments show that entailment-based classifiers assign more accurate pseudo-labels with greater task adaptability and much fewer hyperparameters. We find that our method realizes the benefits of both entailment-based classification and self-training.

Our contributions are as follows:
\begin{enumerate}
    \item We present LIME, a novel framework for weakly-supervised text classification that utilizes textual entailment. LIME surpasses current state-of-the-art weakly-supervised methods in all tested benchmarks.  
    \item We show that self-training with pseudo-labels can mitigate unsolved robustness issues in entailment-based classification \cite{ma-etal-2021-issues}. 
    \item We experimentally confirm that higher confidence in pseudo-labels translates to better classification accuracy in self-training. We also find that a balance between filtering out low-confidence labels and preserving a sizable pseudo-training corpus is important.
\end{enumerate}

\section{Background}

\subsection{Weakly-supervised text classification}
In weakly-supervised text classification, the system is allowed to view the entire test set at evaluation time. Having access to all test data allows novel pre-processing approaches unavailable in traditional text classification, such as preliminary clustering of test samples \cite{mekala-shang-2020-contextualized, wang-etal-2021-x} before attempting final classification. In the process, the system has an opportunity to examine overall characteristics of the test set.    

Existing methods for weakly-supervised text classification focus on effectively leveraging such additional information. The dominant approach involves generating pseudo-data to train a neural text classifier. Most methods assign labels to samples in the test set by identifying operative keywords within the text \cite{meng2018weakly}. They obtain seed words that best represent each category name. Then, each sample in the test set is assigned a label with keywords most relevant to its content.   

Later works improve this pipeline by automatically generating seed words \cite{meng-etal-2020-text} or incorporating pre-trained language models to utilize contextual information of representative keywords \cite{mekala-shang-2020-contextualized}.    

Seed-word-based pseudo-labeling, however, is heavily dependent on the existence of representative seed words in test samples. Seed-word-based matching cannot fully utilize information in contextual language representations, because the classification of each document involves brittle global hyperparameters such as the number of total seed words \cite{meng-etal-2020-text} or word embedding distance \cite{wang-etal-2021-x}.    

In this work, we entirely forgo the seed word generation process during pseudo-labeling. We show that replacing seed-word generation with entailment-based text classification is more reliable and performant for text classification with weak supervision.
\subsection{Entailment based text classification}
Textual entailment \cite{entailment, maccartney-manning-2009-extended} measures the likeliness of a sentence appearing after another. Since entailment is evaluated to a probability value, the task can be extended for use in text classification. In entailment-based text classification, classification is posed as a textual entailment problem: given a test document, the system ranks the probabilities that sentences each containing a possible class label (\textit{hypotheses}) will immediately follow the document text. The class label belonging to the most probable hypothesis is selected as the classification prediction. A hypothesis for topic classification, for example, could be “This text is about \textit{<topic>}”. The flexibility in prompt choices for constructing the hypotheses makes entailment-based classification extremely adaptable to different task types.

Although entailment-based sentence scoring is popular in zero- and few-shot text classification \cite{yin-etal-2019-benchmarking, yin-etal-2020-universal}, the robustness of such approaches has recently been called into question \cite{ma-etal-2021-issues}. Since entailment-based classifiers rely heavily on lexical patterns, a large variance is observed in classification performance across different domains. We find that self-training commonly found in weakly-supervised classification mitigates such robustness issues in entailment-based classification to a large degree.

\section{The LIME Framework}

LIME enhances the two-phase weakly-supervised classification pipeline with an entailment-based pseudo-labeling scheme. 

\begin{table}[h]
 \centering
 \begin{tabular}{ll}
    & Examples \\\midrule
    \textit{Test sample} ($t$) & ``I love the food." \\
    \textit{Class label} ($c$) & ``Positive"\\
    \textit{Verbalizer} & "Positive" $\rightarrow$ ``good"\\
    \textit{Prompt} & "It was <$verbalizer(h_i)$>." \\
    \textit{Hypothesis} ($h$) & "It was good."
    \\\bottomrule
 \end{tabular}
 \caption{Example test sample, class label, verbalizer, prompt, and entailment hypothesis. Converting class labels with a verbalizer is an optional procedure.}
 \label{tab:verbalizer}
\end{table}

\begin{table*}[th]
 \centering
 \begin{tabular}{lcccc}\toprule
    \textbf{Dataset} & \textbf{Type} & \textbf{\# of Classes} & \textbf{Dataset size} & \textbf{Prompt} \\\midrule
    20News    & News topic & 5 & 17,871 & \textit{The text is about <class label>.}\\
    AGNews & News topic  & 4 & 120,000 & \textit{The text is about <class label>.} \\
    Yelp & Restarant review & 2 & 38,000 & \textit{It was good.} / \textit{It was bad.}\\
    DBpedia & Wikipedia topic & 14 & 560,000  & \textit{The text is about <class label>.}
    \\\bottomrule
 \end{tabular}
 \caption{Statistics for benchmark datasets.}
 \label{tab:dataset}
\end{table*}
\subsection{Phase 1: Pseudo-labeling}
Textual entailment evaluates the likeliness of a hypothesis $h$ succeeding some text $t$. 

Given $C = \{c_1, c_2, …, c_n\}$, the set of all possible labels for $t$, we generate $H = \{h_1, h_2, …, h_n\}$, the set of all entailment hypothesis. Every sentence $h_i$ asserts that its corresponding $c_i \in C$ is the correct label for $t$. $h_i$ is constructed from a designated \textit{prompt} and an optional \textit{verbalizer} for each dataset \cite{schick-schutze-2021-exploiting}: 

\begin{equation}
    h_i = prompt(verbalizer(c_i)) \nonumber
\end{equation}  

\textit{Prompts} dictate the wording of the hypotheses, while \textit{verbalizers} convert each class label into a terminology better interpreted by entailment models. Pseudo-label for $t$ is chosen as $c_i$ that corresponds to the pair ($t$, $h_i$) with the highest entailment probability. Table ~\ref{tab:verbalizer} provides examples of verbalizers, prompts, and hypotheses.

\subsection{Phase 2: Self-training}

We adopt a similar self-training approach as existing methods in weakly-supervised text classification. We train a BERT-base model \cite{devlin-etal-2019-bert} with a sequence classification feed-forward layer using pseudo-labels obtained in Phase 1. 

We calculate the prediction confidence for each pseudo-label $c_i$ assigned to $t$. Pseudo-labels under a certain confidence threshold are discarded during the text classifier training phase.  

Confidence of label $c_i$ is defined as the softmax over entailment probabilities of all hypotheses:
\begin{equation}
    Confidence(c_i) = \frac{e^{p_i}}{\sum_{j=1}^n e^{p_{j}}} \nonumber
\end{equation}
where $p_i$ is the entailment probability for the text pair ($t$, $h_i$), obtained from the entailment model.

\section{Experiments}

\subsection{Experimental setting}
In every experiment, we use a publically available BART-large model\footnote{https://huggingface.co/facebook/bart-large-mnli} \cite{lewis-etal-2020-bart} trained on the MultiNLI \cite{williams-etal-2018-broad} dataset as our entailment classifier. We also discard pseudo-labels with confidence under 50\%. Although different thresholds lead to higher final F1 scores, we report results with confidence threshold of 50\% for a fair comparison with previous research.

\subsection{Baselines}
We compare LIME with both entailment-based classification (Phase 1 without self-training) and previous research on weakly-supervised text classification. We also include BERT trained with supervision from original labels as a realistic upper bound for weakly-supervised classification. 

\textbf{WestClass} \cite{meng2018weakly} generates pseudo-documents for each class label. \textbf{ConWea} \cite{mekala-shang-2020-contextualized} utilizes a pre-trained language model to discern keywords that carry different meanings under different contexts. \textbf{LotClass} \cite{meng-etal-2020-text} is a framework for text classification using only label names. \textbf{LotClass} mines a pre-trained BERT model for seed words that are most likely to replace each class name. \textbf{X-Class} \cite{wang-etal-2021-x} is a state-of-the-art weakly-supervised classification system that collects seed words within the test documents instead of external sources. Documents are then grouped with a Gaussian Mixture Model before pseudo-labels are assigned. 

\subsection{Datasets}
We run LIME on standard benchmarks in weakly-supervised classification: \textbf{20News} \cite{LANG1995331}, \textbf{AGNews} \cite{NIPS2015_250cf8b5}, \textbf{Yelp reviews} \cite{NIPS2015_250cf8b5}, and \textbf{DBpedia} \cite{NIPS2015_250cf8b5}. Detailed descriptions of each dataset, along with specific prompts used, are recorded in Table~\ref{tab:dataset}. We notably omit NYT datasets used in \citet{10.1145/3366423.3380278} and \citet{wang-etal-2021-x}, because only pre-processed (all lower-cased, pre-tokenized with a specific tokenizer) versions of the data were available. It is not possible to meaningfully evaluate the pseudo-labeling scheme in LIME if test samples are tokenized by a tokenizer different from that coupled with our entailment model.

\begin{table*}[th]
 \centering
 \begin{tabular}{lcccc}\toprule
    \textbf{Model} & \textbf{20News} & \textbf{AGNews} & \textbf{Yelp} & \textbf{DBpedia} \\\midrule
    Supervised    & 96.45 / 96.42 & 93.99 / 93.99 &  95.70 / 95.70 & 98.96 / 98.96\\
    Entailment classifier    & 67.95 / 67.50  & 79.94 / 79.99  & 94.79 / 94.79 & 80.14 / 79.27 \\
    \cmidrule(lr){1-5}
    WeSTClass    & 71.28 / 69.90 & 82.30 / 82.10 &  81.60 / 81.6 & 81.42 / 81.19\\
    ConWea & 75.73 / 73.26 & 74.60 / 74.20 & 71.40 / 71.20 & N/A\\
    LOTClass & 73.78 / 72.53 & 86.89 / 86.82 & 87.75 / 87.68 &  86.66 / 85.98\\
    X-Class  & 78.62 / 77.76 & 85.74 / 85.66 & 90.00 / 90.00 & 91.32 / 91.17 \\
    LIME  & \textbf{79.74 / 79.56} & \textbf{87.21 / 87.16} & \textbf{95.22 / 95.22} & \textbf{92.19 / 92.20}
    \\\bottomrule
 \end{tabular}
 \caption{Experiment results on 4 classification benchmarks. All reported scores in the form \textit{micro-F1 / macro-F1}. Baselines are quoted from \cite{wang-etal-2021-x}.}
 \label{tab:results}
\end{table*}

\section{Results}
\subsection{Classification performance}
Final classification results are recorded in Table~\ref{tab:results}. LIME outperforms all baselines in terms of micro- and macro-F1 scores, even approaching the supervised baseline in the \textbf{Yelp} dataset. 

We also find that training a new classifier with pseudo-labels (Phase 2 of LIME) does not amplify or propagate errors in incorrect pseudo-labels. The final classifier consistently scores roughly 10 points higher in F1 scores compared to the entailment classifier. Our results confirm findings from previous research that employ self-training to improve classification robustness in low-resource regimes \cite{NEURIPS2020_f23d125d, gowal2021selfsupervised}. 

\subsection{Effect of label confidence thresholds}
\begin{figure}[h]
\includegraphics[width=\columnwidth]{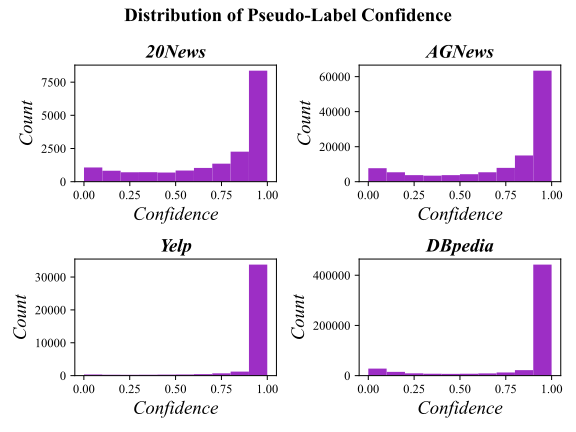}
\caption{Histogram of pseudo-label confidence. More confident pseudo-labels result in more accurate classification self-training.}
\label{fig:confidence}
\end{figure}

Figure~\ref{fig:confidence} plots the spread of pseudo-label confidence produced in Phase 1 of LIME. We confirm that higher average confidence from the entailment classifier in Phase 1 robustly translates to higher classification accuracy for both the entailment classifier and the self-trained classifier in Phase 2.

\begin{figure}[h]
\includegraphics[width=\columnwidth]{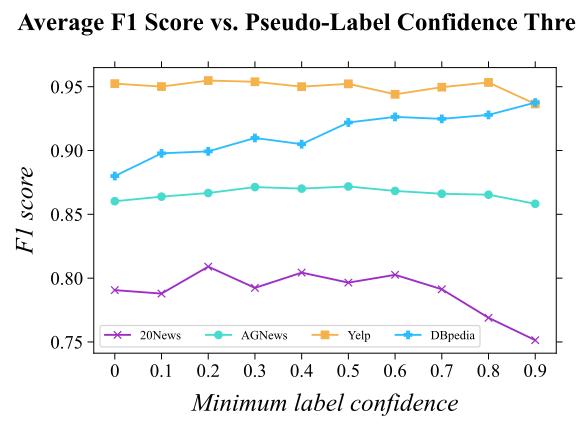}
\caption{Effect of varying confidence thresholds on self-training F1 scores.}
\label{fig:f1}
\end{figure}

Another notable finding is that naively utilizing only high-confidence labels does not always guarantee a more accurate classifier. A trade-off exists between filtering out low-confidence labels and retaining a sizable training corpus. We find that confidence cut-off from 50\% to 70\% strikes a good balance between the two obligations (Figure~\ref{fig:f1}). 

\section{Conclusions}

LIME proposes a streamlined pseudo-labeling method for weakly-supervised text classification. The framework combines flexibility of entailment-based classification with robustness of self-training. The resulting text classifier outperforms previous state-of-the-art in weakly-supervised classification.    
We also investigate the effect of pseudo-label confidence thresholds on self-trained classifier performance. Entailment model confidence accurately reflects label accuracy, but size of the pseudo-training set is also important for robust classification.

We identify several avenues for future research. For a fair comparison with previous research, we did not modify the self-training step with more advanced neural classifier architectures or confidence-aware self-training schemes \cite{NEURIPS2020_f23d125d}. Other auxiliary tasks, such as question-answering \cite{https://doi.org/10.48550/arxiv.1806.08730} or next sentence prediction \cite{ma-etal-2021-issues} can also extend the LIME framework as alternate pseudo-classifiers. 

\bibliography{anthology,custom}

\begin{thebibliography}{19}
\expandafter\ifx\csname natexlab\endcsname\relax\def\natexlab#1{#1}\fi

\bibitem[{Devlin et~al.(2019)Devlin, Chang, Lee, and
  Toutanova}]{devlin-etal-2019-bert}
Jacob Devlin, Ming-Wei Chang, Kenton Lee, and Kristina Toutanova. 2019.
\newblock \href {https://doi.org/10.18653/v1/N19-1423} {{BERT}: Pre-training of
  deep bidirectional transformers for language understanding}.
\newblock In \emph{Proceedings of the 2019 Conference of the North {A}merican
  Chapter of the Association for Computational Linguistics: Human Language
  Technologies, Volume 1 (Long and Short Papers)}, pages 4171--4186,
  Minneapolis, Minnesota. Association for Computational Linguistics.

\bibitem[{Fyodorov et~al.(2000)Fyodorov, Winter, and Francez}]{entailment}
Yaroslav Fyodorov, Yoad Winter, and Nissim Francez. 2000.
\newblock A natural logic inference system.

\bibitem[{Gowal et~al.(2021)Gowal, Huang, van~den Oord, Mann, and
  Kohli}]{gowal2021selfsupervised}
Sven Gowal, Po-Sen Huang, Aaron van~den Oord, Timothy Mann, and Pushmeet Kohli.
  2021.
\newblock \href {https://openreview.net/forum?id=bgQek2O63w} {Self-supervised
  adversarial robustness for the low-label, high-data regime}.
\newblock In \emph{International Conference on Learning Representations}.

\bibitem[{Lang(1995)}]{LANG1995331}
Ken Lang. 1995.
\newblock \href
  {https://doi.org/https://doi.org/10.1016/B978-1-55860-377-6.50048-7}
  {Newsweeder: Learning to filter netnews}.
\newblock In Armand Prieditis and Stuart Russell, editors, \emph{Machine
  Learning Proceedings 1995}, pages 331--339. Morgan Kaufmann, San Francisco
  (CA).

\bibitem[{Lewis et~al.(2020)Lewis, Liu, Goyal, Ghazvininejad, Mohamed, Levy,
  Stoyanov, and Zettlemoyer}]{lewis-etal-2020-bart}
Mike Lewis, Yinhan Liu, Naman Goyal, Marjan Ghazvininejad, Abdelrahman Mohamed,
  Omer Levy, Veselin Stoyanov, and Luke Zettlemoyer. 2020.
\newblock \href {https://doi.org/10.18653/v1/2020.acl-main.703} {{BART}:
  Denoising sequence-to-sequence pre-training for natural language generation,
  translation, and comprehension}.
\newblock In \emph{Proceedings of the 58th Annual Meeting of the Association
  for Computational Linguistics}, pages 7871--7880, Online. Association for
  Computational Linguistics.

\bibitem[{Ma et~al.(2021)Ma, Yao, Lin, and Zhao}]{ma-etal-2021-issues}
Tingting Ma, Jin-Ge Yao, Chin-Yew Lin, and Tiejun Zhao. 2021.
\newblock \href {https://doi.org/10.18653/v1/2021.acl-short.99} {Issues with
  entailment-based zero-shot text classification}.
\newblock In \emph{Proceedings of the 59th Annual Meeting of the Association
  for Computational Linguistics and the 11th International Joint Conference on
  Natural Language Processing (Volume 2: Short Papers)}, pages 786--796,
  Online. Association for Computational Linguistics.

\bibitem[{MacCartney and Manning(2009)}]{maccartney-manning-2009-extended}
Bill MacCartney and Christopher~D. Manning. 2009.
\newblock \href {https://aclanthology.org/W09-3714} {An extended model of
  natural logic}.
\newblock In \emph{Proceedings of the Eight International Conference on
  Computational Semantics}, pages 140--156, Tilburg, The Netherlands.
  Association for Computational Linguistics.

\bibitem[{McCann et~al.(2018)McCann, Keskar, Xiong, and
  Socher}]{https://doi.org/10.48550/arxiv.1806.08730}
Bryan McCann, Nitish~Shirish Keskar, Caiming Xiong, and Richard Socher. 2018.
\newblock \href {https://doi.org/10.48550/ARXIV.1806.08730} {The natural
  language decathlon: Multitask learning as question answering}.

\bibitem[{Mekala and Shang(2020)}]{mekala-shang-2020-contextualized}
Dheeraj Mekala and Jingbo Shang. 2020.
\newblock \href {https://doi.org/10.18653/v1/2020.acl-main.30} {Contextualized
  weak supervision for text classification}.
\newblock In \emph{Proceedings of the 58th Annual Meeting of the Association
  for Computational Linguistics}, pages 323--333, Online. Association for
  Computational Linguistics.

\bibitem[{Meng et~al.(2020{\natexlab{a}})Meng, Huang, Wang, Wang, Zhang, Zhang,
  and Han}]{10.1145/3366423.3380278}
Yu~Meng, Jiaxin Huang, Guangyuan Wang, Zihan Wang, Chao Zhang, Yu~Zhang, and
  Jiawei Han. 2020{\natexlab{a}}.
\newblock \href {https://doi.org/10.1145/3366423.3380278} {\emph{Discriminative
  Topic Mining via Category-Name Guided Text Embedding}}, page 2121–2132.
  Association for Computing Machinery, New York, NY, USA.

\bibitem[{Meng et~al.(2018)Meng, Shen, Zhang, and Han}]{meng2018weakly}
Yu~Meng, Jiaming Shen, Chao Zhang, and Jiawei Han. 2018.
\newblock Weakly-supervised neural text classification.
\newblock In \emph{proceedings of the 27th ACM International Conference on
  information and knowledge management}, pages 983--992.

\bibitem[{Meng et~al.(2020{\natexlab{b}})Meng, Zhang, Huang, Xiong, Ji, Zhang,
  and Han}]{meng-etal-2020-text}
Yu~Meng, Yunyi Zhang, Jiaxin Huang, Chenyan Xiong, Heng Ji, Chao Zhang, and
  Jiawei Han. 2020{\natexlab{b}}.
\newblock \href {https://doi.org/10.18653/v1/2020.emnlp-main.724} {Text
  classification using label names only: A language model self-training
  approach}.
\newblock In \emph{Proceedings of the 2020 Conference on Empirical Methods in
  Natural Language Processing (EMNLP)}, pages 9006--9017, Online. Association
  for Computational Linguistics.

\bibitem[{Mukherjee and Awadallah(2020)}]{NEURIPS2020_f23d125d}
Subhabrata Mukherjee and Ahmed Awadallah. 2020.
\newblock \href
  {https://proceedings.neurips.cc/paper/2020/file/f23d125da1e29e34c552f448610ff25f-Paper.pdf}
  {Uncertainty-aware self-training for few-shot text classification}.
\newblock In \emph{Advances in Neural Information Processing Systems},
  volume~33, pages 21199--21212. Curran Associates, Inc.

\bibitem[{Schick and Sch{\"u}tze(2021)}]{schick-schutze-2021-exploiting}
Timo Schick and Hinrich Sch{\"u}tze. 2021.
\newblock \href {https://doi.org/10.18653/v1/2021.eacl-main.20} {Exploiting
  cloze-questions for few-shot text classification and natural language
  inference}.
\newblock In \emph{Proceedings of the 16th Conference of the European Chapter
  of the Association for Computational Linguistics: Main Volume}, pages
  255--269, Online. Association for Computational Linguistics.

\bibitem[{Wang et~al.(2021)Wang, Mekala, and Shang}]{wang-etal-2021-x}
Zihan Wang, Dheeraj Mekala, and Jingbo Shang. 2021.
\newblock \href {https://doi.org/10.18653/v1/2021.naacl-main.242} {{X}-class:
  Text classification with extremely weak supervision}.
\newblock In \emph{Proceedings of the 2021 Conference of the North American
  Chapter of the Association for Computational Linguistics: Human Language
  Technologies}, pages 3043--3053, Online. Association for Computational
  Linguistics.

\bibitem[{Williams et~al.(2018)Williams, Nangia, and
  Bowman}]{williams-etal-2018-broad}
Adina Williams, Nikita Nangia, and Samuel Bowman. 2018.
\newblock \href {https://doi.org/10.18653/v1/N18-1101} {A broad-coverage
  challenge corpus for sentence understanding through inference}.
\newblock In \emph{Proceedings of the 2018 Conference of the North {A}merican
  Chapter of the Association for Computational Linguistics: Human Language
  Technologies, Volume 1 (Long Papers)}, pages 1112--1122, New Orleans,
  Louisiana. Association for Computational Linguistics.

\bibitem[{Yin et~al.(2019)Yin, Hay, and Roth}]{yin-etal-2019-benchmarking}
Wenpeng Yin, Jamaal Hay, and Dan Roth. 2019.
\newblock \href {https://doi.org/10.18653/v1/D19-1404} {Benchmarking zero-shot
  text classification: Datasets, evaluation and entailment approach}.
\newblock In \emph{Proceedings of the 2019 Conference on Empirical Methods in
  Natural Language Processing and the 9th International Joint Conference on
  Natural Language Processing (EMNLP-IJCNLP)}, pages 3914--3923, Hong Kong,
  China. Association for Computational Linguistics.

\bibitem[{Yin et~al.(2020)Yin, Rajani, Radev, Socher, and
  Xiong}]{yin-etal-2020-universal}
Wenpeng Yin, Nazneen~Fatema Rajani, Dragomir Radev, Richard Socher, and Caiming
  Xiong. 2020.
\newblock \href {https://doi.org/10.18653/v1/2020.emnlp-main.660} {Universal
  natural language processing with limited annotations: Try few-shot textual
  entailment as a start}.
\newblock In \emph{Proceedings of the 2020 Conference on Empirical Methods in
  Natural Language Processing (EMNLP)}, pages 8229--8239, Online. Association
  for Computational Linguistics.

\bibitem[{Zhang et~al.(2015)Zhang, Zhao, and LeCun}]{NIPS2015_250cf8b5}
Xiang Zhang, Junbo Zhao, and Yann LeCun. 2015.
\newblock \href
  {https://proceedings.neurips.cc/paper/2015/file/250cf8b51c773f3f8dc8b4be867a9a02-Paper.pdf}
  {Character-level convolutional networks for text classification}.
\newblock In \emph{Advances in Neural Information Processing Systems},
  volume~28. Curran Associates, Inc.

\end{thebibliography}
\bibliographystyle{acl_natbib}

\end{document}